\documentclass{article}

\usepackage[preprint]{neurips_2026}

\usepackage[utf8]{inputenc} 
\usepackage[T1]{fontenc}    
\usepackage{url}            
\usepackage{booktabs}       
\usepackage{amsfonts}       
\usepackage{nicefrac}       
\usepackage{microtype}      
\usepackage[table,xcdraw]{xcolor}         
\usepackage{amsmath}
\usepackage{wrapfig}
\usepackage{multirow}
\usepackage{multicol}
\definecolor{citecolor}{HTML}{2980b9}
\definecolor{linkcolor}{HTML}{c0392b}
\usepackage[hidelinks,breaklinks=true,colorlinks,bookmarks=false,citecolor=citecolor,linkcolor=linkcolor]{hyperref}
\usepackage{cleveref}
\usepackage{etoc}            
\usepackage{arydshln}  
\usepackage{array}
\usepackage{tcolorbox}
\definecolor{Gray}{gray}{0.93}
\definecolor{MidnightBlue}{RGB}{3,94,146}
\definecolor{Apricot}{RGB}{151,108,170}

\title{What Sketches Tell Us about LVLMs: \\Conventions, Grounding, and Localisation}
\author{%
\href{https://chaitron.github.io/}{\textcolor{citecolor}{Chaitat Utintu}}\textsuperscript{1,2},\quad \href{https://ahmedbourouis.github.io/ahmed-bourouis/}{\textcolor{citecolor}{Ahmed Bourouis}}\textsuperscript{1},\quad \href{https://scholar.google.com/citations?user=7S_jA3IAAAAJ}{\textcolor{citecolor}{Arkaprabha Basu}}\textsuperscript{1},\quad \href{https://www.surrey.ac.uk/people/yi-zhe-song}{\textcolor{citecolor}{Yi-Zhe Song}}\textsuperscript{1,2} \\
  \textsuperscript{1} SketchX, Surrey Institute for People-Centred Artificial Intelligence (PAI), University of Surrey\\
  \textsuperscript{2} UKRI AI: CDT in AI for Digital Media Inclusion\\
  \texttt{\{chaitat.u, a.bourouis, a.basu, y.song\}@surrey.ac.uk} \\
}

\begin{document}

\maketitle
\etocdepthtag.toc{mtmain}

\begin{abstract}

LVLM benchmarks tell us whether a model answers correctly, but not which visual conventions it has retained, nor where those conventions live. Sketches expose this hidden structure. A sketch removes texture, lighting, and photographic context while preserving the strokes humans judged sufficient for recognition. If an LVLM can operate on that abstraction, the surviving signal is structural rather than merely photometric. We introduce a training-free sketch probe with three readouts: temporal part-construction order, spatial attention-based grounding, and the mechanistic layer/head locus of grounding. We evaluate nine LVLMs on part order and five on the attention readouts, over two part-annotated sketch datasets. Three findings emerge. First, LVLM part order tracks human convention: models reproduce canonical orderings where human drawings converge, and become correspondingly variable where human orderings vary. Controls that blank, withhold, and mismatch the sketch locate this convention in the language model. Second, part-level grounding survives the abstraction: attention heads selected from the frozen model with an existing recipe recover part and object regions on object and scene sketches, with no grounding supervision. Third, the grounding signal concentrates in a narrow mid-network band whose location follows the base language model rather than parameter count in the two families we compare, and remains stable across object and scene sketches. The result is a practical diagnostic for sketch-driven LVLMs: no fine-tuning, no auxiliary segmenter, and a layer band that doubles as a head-selection prior, matching or improving grounding mAcc@0.1 in every model and dataset we tested.
\end{abstract}

\begin{figure}[t]
  \centering
  \includegraphics[width=\linewidth]{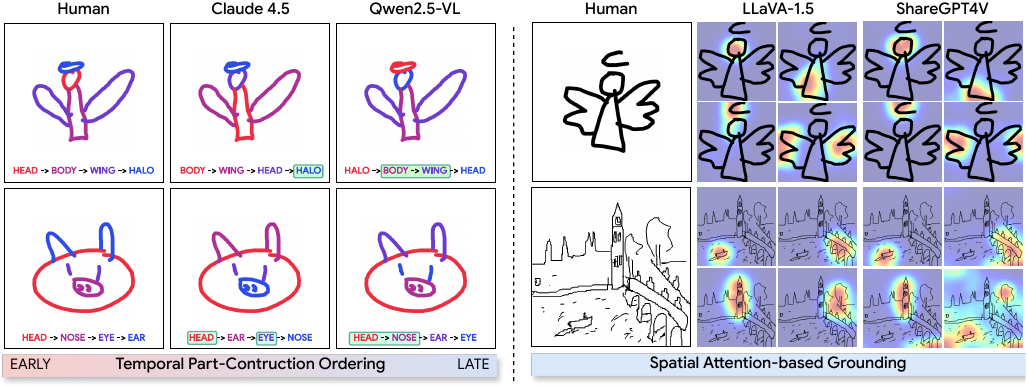}
  \caption{\textbf{Sketches as a probe for LVLMs.} Sketches keep only the strokes humans considered essential, exposing what an LVLM treats as essential. We probe this from three complementary perspectives. \textbf{Left}: \emph{temporal} part-construction order on two object sketches (early-to-late gradient, predicted ordering beneath) against human consensus. \textbf{Right}: \emph{spatial} attention-based grounding on one object and one scene sketch, with per-part / per-object overlays beside the human sketch.}
  \label{fig:teaser}
\end{figure}

\section{Introduction}
\label{sec:intro}

LVLMs can recognise, describe, and even structure familiar objects with striking competence. A coffee cup gets a handle, an opening, and a body in roughly the right arrangement. Standard multimodal benchmarks~\cite{liu2024mmbench,fu2023mme} score the final answer, but leave a central question unresolved: which visual conventions has the model retained, and where in the network are they carried?

This question matters because LVLMs~\cite{liu2024llava,qwen25vl2025,openai2025gpt5,anthropic2025claude45,comanici2025gemini25} are trained on human-produced images and language. Their object knowledge is therefore entangled with the conventions humans use when drawing, depicting, and naming visual structure. A model may answer a part-level question about a photograph correctly by leaning on texture, colour, background, or dataset co-occurrence, without using the part-level schema of the object at all. By a \emph{visual convention} we mean a regularity in how humans depict a concept that is not entailed by its physical structure, such as which parts are drawn and in what order: that a cup has a handle is a fact about cups, whereas drawing the body before the handle is a convention.

Reading these out requires a probe that strips surface signal whilst preserving structural commitments. Natural images are the wrong probe: photographs carry texture, lighting, and contextual scaffolding that the model can lean on, masking whatever schematic structure it has acquired. Colour and texture are correlates of object identity that do not require decomposing an object into parts, whereas spatial arrangement is constitutive of part-level structure, and it is the channel for which SPG~\cite{li2018spg} records human ground truth stroke by stroke. A sketch keeps only the strokes a person judged essential and discards everything else, so what survives is the part-level structural skeleton of the depicted concept. Probing an LVLM with sketches asks the model to operate on this skeleton alone, and what it surfaces is a direct readout of the visual conventions it has retained (Fig.~\ref{fig:teaser}).

\noindent\textbf{What this paper claims.} Such a probe should also read internal state rather than output accuracy. Ours is a training-free diagnostic, deliberately assembled from existing components: shuffled-list ordering for part-construction order, the Localization Heads recipe of Kang et al.~\cite{kang2025localisationheads} for grounding-head selection, and classical sequence metrics~\cite{levenshtein1966edit,hirschberg1975lcs,kendall1938tau} for trajectory comparison. The analytical weight therefore falls on the frozen LVLMs, not on new apparatus.

We instantiate three readouts on two part-annotated sketch datasets~\cite{li2018spg,chowdhury2022fscoco}: temporal part-construction order across nine LVLMs, three of them closed-source, and spatial attention-based grounding and its mechanistic layer/head locus across five open models on two visual stacks. The readouts expose two facts about \emph{what} survives abstraction and one about \emph{where} it is stored:
\begin{itemize}
    \item[(i)] \emph{What conventions persist.} LVLMs reproduce canonical part-orderings on categories where human drawings converge, and become correspondingly unstructured where they do not, deferring to human convention rather than imposing their own schema. Controls that blank, withhold, and mismatch the sketch place this convention in the language model itself.
    \item[(ii)] \emph{What is grounded without training.} Heads selected with the recipe of Kang et al.~\cite{kang2025localisationheads} recover part and object regions on both object and scene sketches from frozen cross-attention alone. Against the trained grounders CLIPSeg~\cite{lueddecke2022clipseg} and GroupViT~\cite{xu2022groupvit}, they overlap the target region more often at coarse and moderate thresholds, so part-level grounding is present in the representation rather than added by fine-tuning. At scale the two readouts couple class by class (LLaVA-1.5-13B Pearson $r=+0.40$).
    \item[(iii)] \emph{Where the apparatus lives.} Grounding heads concentrate in a narrow mid-network band whose location tracks the base language model across the two families we compare (Vicuna at fractional depth $0.35$ to $0.45$, DeepSeek at $0.50$ to $0.65$) and is conserved across object/scene domains ($\Delta = +0.004$). The same heads recur across domains, and ablating the three most frequently selected ones lowers the margin towards the correct answer on a downstream localisation question by $0.118$, against $0.010$ for heads drawn from the same layers.
\end{itemize}

\noindent\textbf{What these insights enable.} Insight \emph{(ii)} is the most immediately actionable: sketch-based retrieval~\cite{bhunia2022sketching}, segmentation, editing, and AR/UI pipelines that currently bolt on a dedicated segmenter could read region grounding directly from an LVLM they already run. Insight \emph{(iii)} has implications for backbone selection in grounding-heavy work: band position follows LM lineage rather than parameter count in the families we compare, so the smaller variant in the right family often suffices, and a band that is causally load-bearing is a natural intervention point for attention-steering methods~\cite{yu2024attentionprompting} that would otherwise have to search the network blind.

\noindent\textbf{Contributions.} We make three contributions. (i) We introduce a training-free sketch probe with temporal, spatial, and mechanistic readouts of LVLM internal organisation, computed from frozen models. (ii) We show that LVLMs track human sketch conventions rather than imposing a fixed model schema, with controls that locate this convention in the language model, and that selected attention heads recover part and object regions on object and scene sketches without grounding supervision. (iii) We localise the grounding signal to a narrow family-dependent mid-network band, confirm by ablation that the selected heads carry it, and show that restricting head search to this band acts as a usable prior, matching or improving grounding mAcc@0.1 across every model and dataset tested.


\begin{figure}[t]
  \centering
    \includegraphics[width=\linewidth]{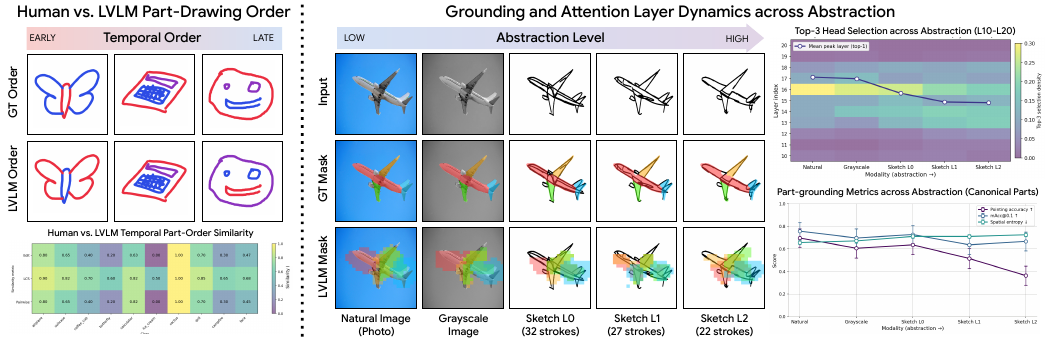}
  \caption{\textbf{Pilot: part-construction ordering and grounding across abstraction.} \textbf{Left}: human (GT) and
LVLM part-drawing trajectories on three SPG~\cite{li2018spg} categories (early$\rightarrow$late
coloured), with mean per-sketch part-order similarity ($\pm$ SD, $N=100$) across ten categories.
\textbf{Right}: LVLM part-localisation on Pascal-Part~\cite{chen2014pascalpart} across
ControlSketch~\cite{arar2025swiftsketch} abstraction levels, top-3 grounding-head selection
(peak shifts to earlier layers), and degrading pointing accuracy and mAcc@0.1 with stable
spatial entropy. See Sec.~\ref{sec:method_motivation}.}
  \label{fig:pilot}
\end{figure}

\vspace{-0.3cm}
\section{Related Work}
\vspace{-0.3cm}
\paragraph{Interpretability in Large Vision-Language Models.}
Visual model interpretability has evolved from CNN saliency~\cite{zeiler2014visualizing,zhou2016cam,selvaraju2017gradcam} through transformer attribution~\cite{abnar2020rollout,chefer2021transformer,aflalo2022vltransparent,salin2022vlprobing} to self-supervised ViTs~\cite{dosovitskiy2021vit} and contrastive VL models~\cite{radford2021clip}. With instruction-tuned LVLMs~\cite{liu2023llava,liu2024llava,bai2023qwenvl,li2023blip2}, grounding can be generated via location tokens~\cite{peng2024kosmos2}, unified with referring~\cite{you2024ferret}, or elicited from LLM-style decoders~\cite{rasheed2024glamm}, while specific attention heads exhibit informative direction~\cite{bousselham2023groundingeverything,kang2025localisationheads} and visual prompts can steer prediction~\cite{yu2024attentionprompting,shtedritski2023redcircle}. Object information stays spatially localised across visual tokens, progressively aligning to interpretable vocabulary~\cite{neo2025interpreting}. Yet these studies probe natural images, where CLIP-based encoders harbour systematic blind spots~\cite{tong2024eyeswideopen}. The word grounding also covers three distinct mechanisms in this literature: generative grounding, where the model emits location tokens or referring expressions~\cite{peng2024kosmos2,you2024ferret}, gradient and attribution saliency~\cite{chefer2021transformer,selvaraju2017gradcam,zeiler2014visualizing,zhou2016cam}, and attention-based localisation, where the region is read directly from attention weights~\cite{bousselham2023groundingeverything,kang2025localisationheads}. Ours is strictly the third, with no coordinate decoding, no gradients, and no auxiliary segmenter. We use sketches as a controlled abstract input, asking whether attention-based localisation survives when texture, lighting, and context are absent, and reporting both its accuracy and the layer band that carries it.

\vspace{-0.2cm}
\paragraph{Sketch understanding and interpretability.}
Sketches are a sparse, abstract medium long used for recognition~\cite{eitz2012tuberlin,ha2017sketchrnn,ribeiro2020sketchformer}, retrieval~\cite{bhunia2022sketching,sain2023clip}, grouping~\cite{qi2022onesketch,li2018spg}, and generation~\cite{wang2023sketchknitter,ge2021creativesketch}. Optimisation-based methods produce object~\cite{vinker2022clipasso} and scene sketches~\cite{vinker2023clipascene}, and benchmarks~\cite{bourouis2024scenesketch,chowdhury2022fscoco,chowdhury2023scenetrilogy} evaluate LVLMs on line drawings, but treat sketches as a domain shift rather than a probe of internal representation. Sketch-specific explainability~\cite{qu2023sketchxai}, saliency~\cite{bhunia2023sketch2saliency}, and detection~\cite{chowdhury2023detection} exploit this further, while cognitive studies use sketches as a window onto visual concept organisation~\cite{fan2020pragmatic,long2024parallel,mukherjee2024seva}, recovering trajectories that reveal stable construction orders~\cite{fan2023drawings}. How LVLMs internalise comparable structure remains untested. We treat the sketch itself as the experimental probe, turning sketches from a domain to evaluate \emph{on} into a probe to evaluate \emph{with}.


\section{Method}
\label{sec:method}
In this section, we first present a \textbf{pilot study} in Sec.~\ref{sec:method_motivation}
examining part-construction ordering and grounding across visual abstraction, motivating
attention as the central probe signal. We then formalise our \textbf{three-perspective
probing framework} in Sec.~\ref{sec:method_overview}, treating the temporal, spatial, and
mechanistic axes as complementary readouts of internal representation. Finally,
Sec.~\ref{sec:method_main} instantiates each perspective as a training-free probe over
object and scene sketches.

\subsection{Pilot study: part-construction ordering and grounding under abstraction}
\label{sec:method_motivation}
Before instantiating our framework at scale, we conduct a pilot study to motivate its central claim: that the part-level structure an LVLM has internalised is revealed most sharply under abstraction, where stripping texture and context exposes its structural commitments. Fig.~\ref{fig:pilot} summarises the findings. \emph{Left (temporal):} we contrast human part-drawing trajectories from SPG~\cite{li2018spg} with LVLM-induced orderings on the same sketches. Both follow coarse-to-fine sequences, and the heatmap reports per-sketch Edit, LCS, and Pairwise similarity (mean $\pm$ SD over 100 sketches per category) across the ten categories whose human drawings most consistently follow a shared canonical order. Agreement is strong on canonical categories such as \emph{calculator} and \emph{coffee\_cup} but degrades on parts where humans themselves disagree, and within-class variance surfaces categories where the LVLM produces near-random orderings rather than a single canonical guess. \emph{Right (spatial \& mechanistic):} we probe attention-based part grounding on Pascal-Part~\cite{chen2014pascalpart} across natural photo, grayscale, and three ControlSketch~\cite{arar2025swiftsketch} abstraction levels. This is a graded ablation over the surface channels: grayscale removes colour while keeping texture and context, and the sketch levels then remove texture and context progressively. LVLM masks track ground truth at low abstraction but degrade as input becomes sparser. The peak grounding layer migrates earlier with increasing abstraction, and pointing accuracy and mAcc@0.1 decline substantially whilst spatial entropy rises only modestly. The model becomes incorrect about \emph{where} parts lie whilst its attention shape changes only slightly. Abstraction therefore selectively erodes spatial accuracy whilst preserving temporal structure and broad attention shape, motivating the probes formalised in Sec.~\ref{sec:method_main}.

\subsection{Problem formulation and method overview}
\label{sec:method_overview}
\paragraph{Problem.}
We treat the sketch input as a controlled stripping of texture, lighting, and surface context that exposes what an LVLM has internalised about objects and their parts. Consider a pre-trained LVLM with $L$ transformer layers and $H$ heads per layer. The input is a pair $(I, p)$, where $I$ is a sketch (object or scene) and $p$ identifies a target part or whole object. A prompt template $T(p)$ (Sec.~\ref{sec:method_main}) converts $p$ into a textual probe query, after which we read out the cross-attention $\alpha_{\ell,h}(p) \in \mathbb{R}^{P^2}$ at each layer $\ell$ and head $h$, where $P^2$ is the number of image patches. The model is used purely via forward passes, with no fine-tuning, auxiliary networks, or external segmentation. Our goal is to characterise the part-level structure that persists in the LVLM's internal state under sketch input. The pilot study (Sec.~\ref{sec:method_motivation}) reveals three signals to formalise: \emph{temporal} part-construction order, \emph{spatial} attention-based grounding, and \emph{mechanistic} layer/head locus.

\begin{figure}[t]
  \centering
  \includegraphics[width=\linewidth]{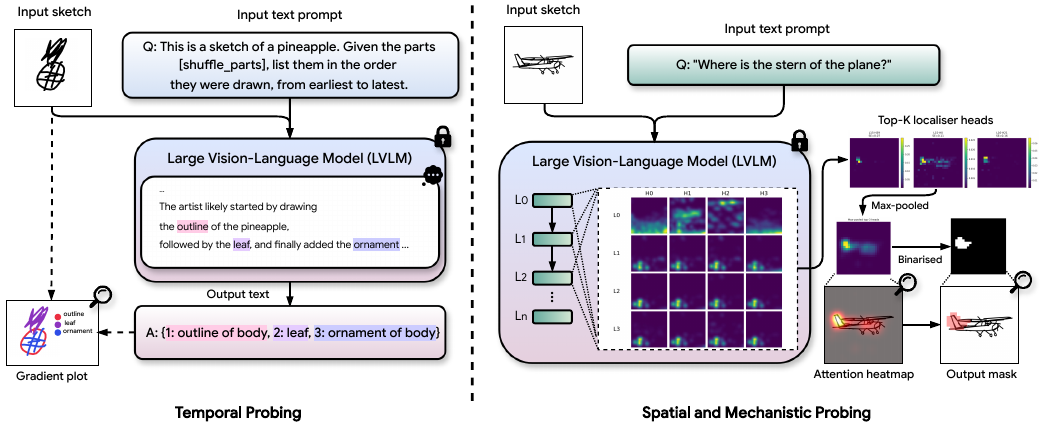}
  \caption{\textbf{Three-perspective probing framework.} The same frozen LVLM is queried twice on a single sketch under two protocols. \textbf{Left (temporal probing):} the model receives the sketch with a shuffled set of parts and is asked to return them in drawing order. \textbf{Right (spatial and mechanistic probing):} the model is asked to locate a named part, and attention is filtered to the top-$k$ grounding heads and max-pooled into a grounding mask.}
  \label{fig:method}
\end{figure}

\paragraph{Three readouts.}
Fig.~\ref{fig:method} illustrates the framework. Rather than a single aggregate score, it returns three readouts of internal state from forward passes of one frozen LVLM on $(I, p)$. The \emph{temporal} readout records the order $\pi(p)$ in which the model returns a shuffled part list and compares it with human drawing trajectories by edit distance, longest common subsequence, and pairwise-order agreement. The \emph{spatial} readout selects a small set $\mathcal{G}$ of grounding heads by image-attention mass and spatial concentration, and turns their combined attention into a part mask. The \emph{mechanistic} readout characterises $\mathcal{G}$ itself, recording the layer and head of each selected head and the spatial entropy of the combined map, and asks whether the band that carries grounding is conserved across models and across the object and scene divide. Sec.~\ref{sec:method_main} defines each in turn.

\newpage
\subsection{Sketch-based probes of part-construction ordering and grounding}
\label{sec:method_main}
\paragraph{Temporal probing.}
We elicit a part-construction trajectory $\pi(p) = (p_1, \dots, p_K)$ in a single forward pass: the model is shown the sketch, the object name, and the part list shuffled into set notation $\{p_1, \dots, p_K\}$, and is asked to return the parts in the order they were drawn. The protocol is purely textual on the parts side: no sketch is re-rendered, so the resulting sequence reflects only the model's internal ordering rather than feedback from a partial drawing. Sec.~\ref{sec:main_results} tests whether the sketch image itself drives this ordering. We compare $\pi(p)$ against human trajectories
$\pi^{\text{human}}$ from SPG~\cite{li2018spg}, which records per-sketch part
strokes and yields a stable modal trajectory per category, using edit
distance, longest common subsequence, and pairwise-order agreement (Kendall's
$\tau$~\cite{kendall1938tau}), which jointly capture absolute position, the longest shared ordered subsequence,
and the relative precedence of part pairs. The parts vocabulary and prompt
template are held fixed across all sketches, so any variation in $\pi(p)$ is
attributable to the model. The temporal probe is restricted to objects, since
SPG supplies human orderings only at the part level within a single object.

\paragraph{Spatial and mechanistic probing.}
We query each input with a template $T(p)$, i.e., \emph{``Where is the \{p\} of the \{c\}?''} for object parts on SPG~\cite{li2018spg} and \emph{``Where is the \{c\}?''} for whole objects on FS-COCO~\cite{chowdhury2022fscoco}, and extract cross-attention from the \emph{last input token}, which under causal masking aggregates information across the whole prompt. Let $\alpha_{\ell,h}(p) \in \mathbb{R}^{P^2}$ denote this attention vector at layer $\ell$, head $h$, restricted to the $P^2$ image patches. Following Kang et al.~\cite{kang2025localisationheads}, who validated the recipe on ten LVLMs including all five used here, we score each head by image-attention mass $\rho_{\ell,h}(p) = \sum_{i \in \mathcal{V}} \alpha_{\ell,h,i}(p)$ over the visual tokens $\mathcal{V}$ and spatial concentration, measured by the spatial entropy $\eta_{\ell,h}(p)$ of the attention map (lower is more concentrated):
\begin{equation}
\eta_{\ell,h}(p) = -\!\sum_{i=1}^{P^2}
   \tilde{\alpha}_{\ell,h,i}(p) \log \tilde{\alpha}_{\ell,h,i}(p),
\;\;
\tilde{\alpha}_{\ell,h}(p) = \alpha_{\ell,h}(p) \,\big/\, \textstyle\sum_i \alpha_{\ell,h,i}(p).
\label{eq:spatial-entropy}
\end{equation}
Heads are scored jointly by these two criteria, reduced via the chord-elbow rule of~\cite{kang2025localisationheads}, and the top $k\!=\!3$ form the grounding-head set $\mathcal{G} \subset \{1, \ldots, L\} \times \{1, \ldots, H\}$. Selected heads are smoothed (mean-subtract, ReLU, Gaussian), combined by element-wise maximum, and thresholded at the 80th percentile to yield a mask $M(p)$, scored against ground-truth part masks (SPG) and object masks (FS-COCO) via pointing accuracy, mAcc@0.1, and spatial entropy (SE). The mechanistic readout reuses the same machinery: we record the (layer, head) coordinates of $\mathcal{G}$ and the per-sample SE, and ask how their distributions vary across models and across the object/scene divide. All three probes require no fine-tuning, no auxiliary heads, and no rendering between iterations.

\newpage
\section{Experiments}
\label{sec:exp}

\subsection{Model sets, protocols, and why they differ across probes}
\label{sec:exp_details}
The three probes draw on overlapping but distinct model sets, dictated by the
methodology each requires. The temporal probe needs only generated text, so it
also covers closed-source models, whereas the attention readouts need access to
the attention weights. For temporal probing
(Tab.~\ref{tab:part-ordering}) we use six open-source models spanning four
families (LLaVA-1.5-7B/13B~\cite{liu2024llava},
Qwen2.5-VL-3B/7B~\cite{qwen25vl2025}, Molmo-7B-D~\cite{deitke2025molmo},
InternVL2-8B~\cite{chen2024internvl}) and three closed-source models from
the smallest tier of each frontier family (GPT-5 (nano)~\cite{openai2025gpt5},
Claude Haiku 4.5~\cite{anthropic2025claude45},
Gemini 2.5 Flash-Lite~\cite{comanici2025gemini25}). All models use deterministic
decoding (temperature $0$), and closed-source predictions are averaged over three runs per
item for stability. Trajectories are elicited on SPG~\cite{li2018spg}
(25 categories) under a class-consensus protocol: for each category we identify
the modal human trajectory and sample 100 strict-mode-matching sketches,
presenting them to each LVLM with the part list shuffled and displayed in
set notation $\{p_1,\dots,p_K\}$ to neutralise input ordering bias. Per-class
scores aggregate over the 100 sketches, and we report mean $\pm$ SD across the 25
categories to expose inter-class variance, which is itself a finding.
For attention-based grounding (Tab.~\ref{tab:grounding}) and the mechanistic
profile (Fig.~\ref{fig:mechanistic}) we use five open models.
LLaVA-1.5-7B/13B and ShareGPT4V-7B~\cite{chen2024sharegpt4v} use a CLIP
ViT-L/14 vision encoder and a Vicuna language model, whereas
DeepSeek-VL-1.3B/7B~\cite{lu2024deepseekvl} pair a SigLIP-based vision encoder
and their own image processor with DeepSeek-LLM. The set therefore spans two
distinct visual stacks and two base language model families, and the
Localization Heads recipe~\cite{kang2025localisationheads} transfers to both
unchanged.
Cross-attention $\alpha_{\ell,h}(p)$ from the last input token to the
$P^2$ image patches is extracted across all $L$ layers and $H$ heads, smoothed
per head (mean-subtract, ReLU, Gaussian $\sigma=1$, normalise), and aggregated
over the top-$k\!=\!3$ heads by image-attention mass and spatial concentration. Grounding is evaluated on SPG (object) and
FS-COCO~\cite{chowdhury2022fscoco} (scene), with FS-COCO scene-level segmentation masks taken from FS-COCO-Seg~\cite{bourouis2024scenesketch},  against CLIPSeg~\cite{lueddecke2022clipseg}
and GroupViT~\cite{xu2022groupvit}, with dataset-level means reported from a
single forward pass with no fine-tuning.

\subsection{Main results}
\label{sec:main_results}

\begin{wrapfigure}{r}{0.4\textwidth}
    \centering
    \vspace{-5mm}
    \includegraphics[width=0.4\textwidth, height=7cm]{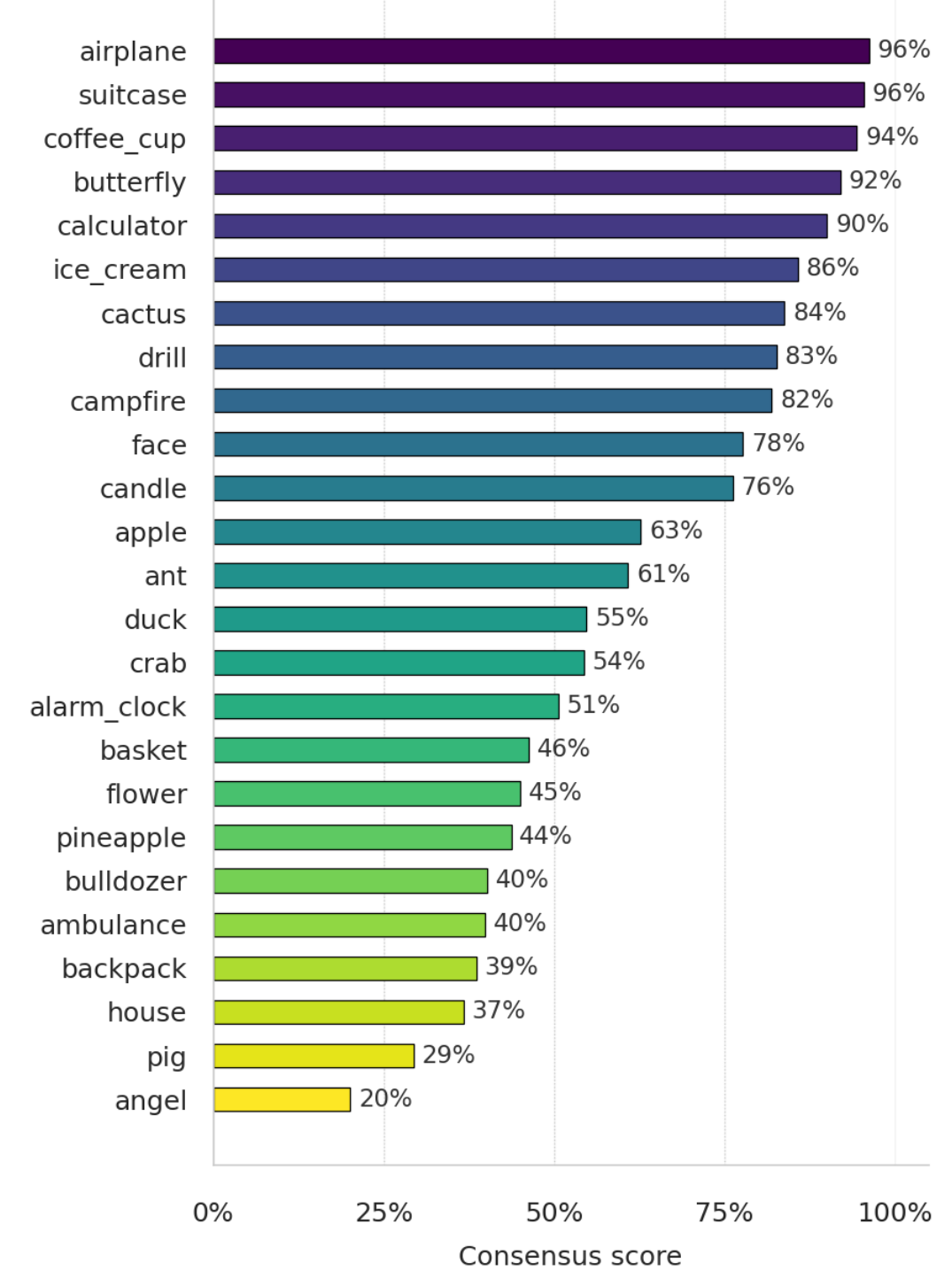}
    \vspace{-6mm}
    \caption{Human consensus statistics across all SPG~\cite{li2018spg} sketch categories.}
    \label{fig:spg_consensus}
    \vspace{-4mm}
\end{wrapfigure}

\begin{figure}[t]
    \centering
    \includegraphics[width=.85\linewidth]{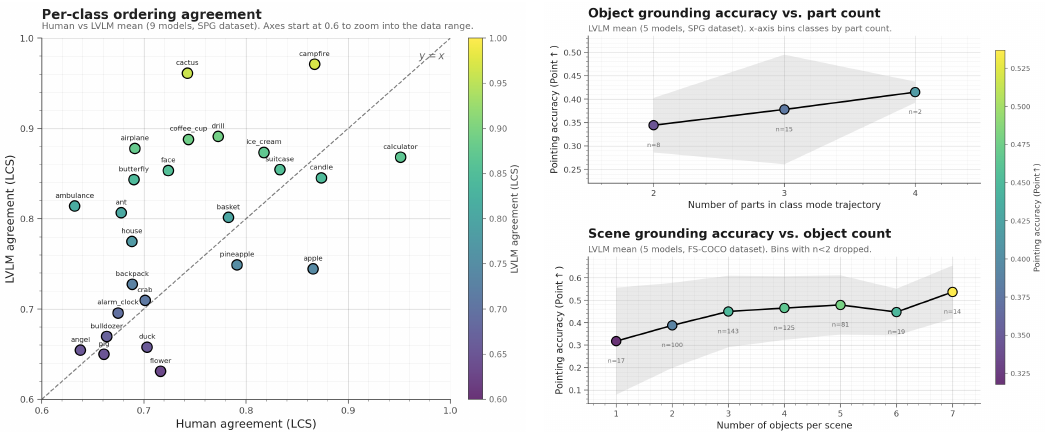}
    \caption{\textbf{Summary of ordering and grounding behaviour.}
    \textbf{Left}: per-class human-LVLM ordering agreement under
    LCS~\cite{hirschberg1975lcs}, with viridis colour encoding LVLM
    agreement. \textbf{Right}: LVLM grounding accuracy as a function of
    input complexity, on object (SPG, parts) and scene (FS-COCO, objects).}
    \label{fig:summary}
\end{figure}

\textbf{Part-construction order.} Before contrasting models, we examine
the human side of SPG. Fig.~\ref{fig:spg_consensus} ranks all 25
categories by strict-mode consensus, and reveals a strongly bimodal
distribution: canonical objects such as \emph{airplane}, \emph{suitcase},
and \emph{coffee\_cup} elicit near-unanimous part orders ($\geq$~90\%),
whereas creature-like or symmetry-rich categories such as \emph{angel},
\emph{pig}, and \emph{house} fall below 40\%, indicating that humans
themselves do not converge on a single drawing schema. This dataset-side
heterogeneity sets the bar against which any LVLM prediction must be
read. Tab.~\ref{tab:part-ordering} reports the resulting agreement under
all three metrics. The closed-source row leads as expected, exceeding
every open-source model. Within the open-source families, however, the
gap is markedly narrower than on standard multimodal benchmarks:
Qwen2.5-VL-7B reaches $0.84$ LCS, within $0.05$ of the closed-source
leader, suggesting that part-construction order is a relatively shallow
inference once a model has the object's name and its part list.
Newer open models narrow the gap further (App.~\ref{app:qwen3}). The per-class
view in Fig.~\ref{fig:summary} (left) reveals where the alignment
holds: categories with strong human consensus (\emph{calculator},
\emph{campfire}, \emph{ice\_cream}) cluster near the diagonal with
high LVLM agreement, whereas categories where humans themselves
disagree (\emph{angel}, \emph{pig}, \emph{flower}) drop into the
lower-left, with LVLM agreement degrading commensurately. LVLMs
internalise the canonical part schemas humans use when those schemas
are stable. Where human consensus dissolves, model behaviour becomes
correspondingly unstructured rather than collapsing onto a single canonical guess.
Across the 25 categories, per-category agreement correlates with human consensus
(mean Pearson $r=+0.58$ over all nine models, significant in eight, App.~\ref{app:consensus}).
Memorisation of SPG would not produce this pattern, since a stored
ordering would be reproduced equally well where humans disagree.

\textbf{Does the sketch drive the ordering?} As the prompt supplies the object name
and its parts as text, we test what the image adds. With prompt, parts, shuffle, and decoding fixed, we replace the
sketch with a blank canvas, remove it (Tab.~\ref{tab:controls}), or substitute a
sketch of another category. Of eighteen paired comparisons over the 25 categories
(two models, three conditions, three metrics), none favours the real sketch, and
the only significant one favours the no-image condition (Pair, $p=0.04$). A test on
sketches whose human order departs from the canonical one ($p=0.26$ under
clustering by category) and a forced choice with chance at 50\% by construction
agree (App.~\ref{app:controls}). The ordering convention is therefore carried by
the language model. SPG remains the instrument that measures it, and the spatial
and mechanistic readouts, which read attention over image patches, are unaffected.

\begin{table}[t]
    \small
    \centering
    \setlength{\tabcolsep}{15pt}
    \caption{\textbf{Part-construction order on object sketches.} LVLM trajectories $\pi(p)$ compared against human trajectories $\pi^{\text{human}}$ on SPG~\cite{li2018spg} under the class-consensus protocol, evaluated with edit distance~\cite{levenshtein1966edit}, longest common subsequence~\cite{hirschberg1975lcs}, and pairwise-order agreement (Kendall's $\tau$)~\cite{kendall1938tau}. Best per column is highlighted within closed-source (\colorbox{Apricot!12}{purple}) and open-source (\colorbox{MidnightBlue!12}{blue}).}
    \vspace{-2mm}
    \label{tab:part-ordering}
    \resizebox{\linewidth}{!}{
    \begin{tabular}{lccc}
    \toprule
    \multirow{2}{*}{\textbf{Model}} & \multicolumn{3}{c}{\textbf{SPG~\cite{li2018spg}}} \\
    \cmidrule(lr){2-4}
    & \textbf{Edit$\uparrow$} & \textbf{LCS$\uparrow$} & \textbf{Pair$\uparrow$} \\
    \midrule
    \multicolumn{4}{l}{\textbf{\textit{Closed-source LVLMs}}} \\
    
    GPT-5 (nano)~\cite{openai2025gpt5}                         & 0.72 {\scriptsize($\pm$0.26)} & 0.85 {\scriptsize($\pm$0.14)} & 0.81 {\scriptsize($\pm$0.19)} \\
    
    \cellcolor{Apricot!12}\textbf{Claude Haiku 4.5~\cite{anthropic2025claude45}}        & \cellcolor{Apricot!12}\textbf{0.80 {\scriptsize($\pm$0.20)}}   & \cellcolor{Apricot!12}\textbf{0.89 {\scriptsize($\pm$0.11)}} & \cellcolor{Apricot!12}\textbf{0.88 {\scriptsize($\pm$0.13)}} \\
    
    Gemini 2.5 Flash-Lite~\cite{comanici2025gemini25}            & 0.73 {\scriptsize($\pm$0.25)} & 0.85 {\scriptsize($\pm$0.14)} & 0.83 {\scriptsize($\pm$0.17)} \\
    \midrule
    \multicolumn{4}{l}{\textbf{\textit{Open-source LVLMs}}} \\
    
    InternVL2-8B~\cite{chen2024internvl}                & 0.66 {\scriptsize($\pm$0.28)} & 0.80 {\scriptsize($\pm$0.18)} & 0.72 {\scriptsize($\pm$0.28)} \\
    
    Molmo-7B-D~\cite{deitke2025molmo}                   & 0.63 {\scriptsize($\pm$0.24)} & 0.78 {\scriptsize($\pm$0.16)} & 0.71 {\scriptsize($\pm$0.22)} \\
    
    Qwen2.5-VL-3B~\cite{qwen25vl2025}                   & 0.57 {\scriptsize($\pm$0.26)} & 0.77 {\scriptsize($\pm$0.14)} & 0.64 {\scriptsize($\pm$0.24)} \\
    
    \cellcolor{MidnightBlue!12}\textbf{Qwen2.5-VL-7B~\cite{qwen25vl2025}}                   & \cellcolor{MidnightBlue!12}\textbf{0.69 {\scriptsize($\pm$0.22)}}   & \cellcolor{MidnightBlue!12}\textbf{0.84 {\scriptsize($\pm$0.12)}} & \cellcolor{MidnightBlue!12}\textbf{0.80 {\scriptsize($\pm$0.15)}} \\
    
    LLaVA-1.5-7B~\cite{liu2024llava}                    & 0.38 {\scriptsize($\pm$0.24)} & 0.66 {\scriptsize($\pm$0.12)} & 0.46 {\scriptsize($\pm$0.26)} \\
    
    LLaVA-1.5-13B~\cite{liu2024llava}                   & 0.46 {\scriptsize($\pm$0.16)} & 0.69 {\scriptsize($\pm$0.09)} & 0.51 {\scriptsize($\pm$0.18)} \\
    \bottomrule
    \end{tabular}
    }
\end{table}

\begin{table}[t]
    \small
    \centering
    \setlength{\tabcolsep}{5pt}
    \caption{\textbf{Attention-based grounding on object and scene sketches.} Cross-attention grounding accuracy of LVLM grounding heads on object (SPG) and scene (FS-COCO) sketches, under pointing accuracy~\cite{zhang2016pointing}, mAcc@0.1~\cite{bilen2016wsddn}, and spatial entropy~\cite{riche2013saliency}. Best per column is highlighted within trained grounders (\colorbox{Apricot!12}{purple}, shown for reference) and LVLMs (\colorbox{MidnightBlue!12}{blue}).}
    \vspace{-1mm}
    \label{tab:grounding}
    \resizebox{\linewidth}{!}{
    \begin{tabular}{l p{10pt} ccc ccc}
    \toprule
    \multirow{2}{*}{\textbf{Model}}
    &
    & \multicolumn{3}{c}{\textbf{Object, SPG~\cite{li2018spg}}}
    & \multicolumn{3}{c}{\textbf{Scene, FS-COCO~\cite{chowdhury2022fscoco}}} \\
    \cmidrule(lr){3-5} \cmidrule(lr){6-8}
    &
    & \textbf{Point$\uparrow$}
    & \textbf{mAcc@0.1$\uparrow$}
    & \textbf{SE$\downarrow$}
    & \textbf{Point$\uparrow$}
    & \textbf{mAcc@0.1$\uparrow$}
    & \textbf{SE$\downarrow$} \\
    \midrule

    \multicolumn{8}{l}{\textbf{\textit{Trained grounders}}} \\

    \cellcolor{Apricot!12}\textbf{CLIPSeg~\cite{lueddecke2022clipseg}}
    & \cellcolor{Apricot!12}\textbf{} & \cellcolor{Apricot!12}\textbf{0.56} & 0.27 & \cellcolor{Apricot!12}\textbf{0.96} & \cellcolor{Apricot!12}\textbf{0.67} & 0.40 & \cellcolor{Apricot!12}\textbf{0.96} \\

    \cellcolor{Apricot!12}\textbf{GroupViT~\cite{xu2022groupvit}}
    & \cellcolor{Apricot!12}\textbf{} & 0.34 & \cellcolor{Apricot!12}\textbf{0.59} & 0.98 & 0.39 & \cellcolor{Apricot!12}\textbf{0.60} & 0.97 \\

    \midrule
    \multicolumn{8}{l}{\textbf{\textit{Training-free LVLM probes}}} \\

    ShareGPT4V-7B~\cite{chen2024sharegpt4v}
    & & 0.42 & 0.72 & 0.87 & 0.45 & 0.72 & 0.86 \\

    DeepSeek-VL-1.3B~\cite{lu2024deepseekvl}
    & & 0.31 & 0.56 & 0.92 & 0.34 & 0.61 & 0.91 \\

    DeepSeek-VL-7B~\cite{lu2024deepseekvl}
    & & 0.23 & 0.48 & 0.85 & 0.33 & 0.65 & 0.84 \\

    LLaVA-1.5-7B~\cite{liu2024llava}
    & & 0.43 & 0.65 & 0.88 & 0.53 & 0.69 & 0.87 \\

    \rowcolor{MidnightBlue!12}
    \textbf{LLaVA-1.5-13B~\cite{liu2024llava}}
    & & \textbf{0.46} & \textbf{0.73} & \textbf{0.84} & \textbf{0.61} & \textbf{0.78} & \textbf{0.83} \\

    \bottomrule
    \end{tabular}
    }
\end{table}

\begin{table}[t]
  \centering
  \small
  \begin{minipage}[t]{0.50\linewidth}
    \centering
    \caption{\textbf{Image controls on the temporal probe.} 50 sketches per category, prompt and shuffles fixed. Real-sketch rows are within $0.01$ of Tab.~\ref{tab:part-ordering}.}
    \label{tab:controls}
    \setlength{\tabcolsep}{3pt}
    \resizebox{\linewidth}{!}{
    \begin{tabular}{lccc ccc}
    \toprule
    & \multicolumn{3}{c}{\textbf{Qwen2.5-VL-7B}} & \multicolumn{3}{c}{\textbf{LLaVA-1.5-7B}} \\
    \cmidrule(lr){2-4}\cmidrule(lr){5-7}
    \textbf{Input} & \textbf{Edit$\uparrow$} & \textbf{LCS$\uparrow$} & \textbf{Pair$\uparrow$} & \textbf{Edit$\uparrow$} & \textbf{LCS$\uparrow$} & \textbf{Pair$\uparrow$} \\
    \midrule
    Real sketch  & 0.69 & 0.84 & 0.79 & 0.37 & 0.66 & 0.46 \\
    Blank canvas & 0.69 & 0.84 & 0.79 & 0.39 & 0.68 & 0.50 \\
    No image     & 0.72 & 0.86 & 0.83 & 0.44 & 0.69 & 0.51 \\
    \bottomrule
    \end{tabular}}
  \end{minipage}%
  \hfill
  \begin{minipage}[t]{0.46\linewidth}
    \centering
    \caption{\textbf{Grounding at stricter overlap.} All methods re-scored at IoU $0.3$ and $0.5$. mAcc@0.1 is in Tab.~\ref{tab:grounding}.}
    \label{tab:strict}
    \setlength{\tabcolsep}{3pt}
    \resizebox{\linewidth}{!}{
    \begin{tabular}{lcc cc}
    \toprule
    & \multicolumn{2}{c}{\textbf{SPG}} & \multicolumn{2}{c}{\textbf{FS-COCO}} \\
    \cmidrule(lr){2-3}\cmidrule(lr){4-5}
    \textbf{Model} & \textbf{@0.3$\uparrow$} & \textbf{@0.5$\uparrow$} & \textbf{@0.3$\uparrow$} & \textbf{@0.5$\uparrow$} \\
    \midrule
    CLIPSeg~\cite{lueddecke2022clipseg} & 0.17 & 0.12 & 0.31 & 0.20 \\
    GroupViT~\cite{xu2022groupvit}      & 0.28 & 0.11 & 0.21 & 0.08 \\
    LLaVA-1.5-13B                        & 0.29 & 0.06 & 0.35 & 0.11 \\
    \bottomrule
    \end{tabular}}
  \end{minipage}
\end{table}

\textbf{Attention-based grounding.}
We turn from temporal ordering to where the LVLM places its attention.
Tab.~\ref{tab:grounding} reports cross-attention grounding accuracy on
object (SPG) and scene (FS-COCO) sketches under pointing accuracy,
mAcc@0.1, and spatial entropy. The strongest LVLM probe, LLaVA-1.5-13B,
reaches $0.46$ Point and $0.73$ mAcc@0.1 on SPG, and $0.61$ Point and
$0.78$ mAcc@0.1 on FS-COCO. The comparison against the trained
grounders depends on what is measured. LLaVA-1.5-13B exceeds
GroupViT~\cite{xu2022groupvit} on Point but trails
CLIPSeg~\cite{lueddecke2022clipseg} ($0.46$ vs $0.56$ on SPG,
$0.61$ vs $0.67$ on FS-COCO), whereas on mAcc@0.1, which asks only whether
the predicted mask overlaps the target region, it leads both, with CLIPSeg at
$0.40$ on FS-COCO against $0.78$. At stricter thresholds (Tab.~\ref{tab:strict})
the lead holds at mAcc@0.3, significant in three of four comparisons, and is gone
at mAcc@0.5, where the trained grounders are nominally ahead in three of four
comparisons, none significant. Frozen cross-attention therefore matches or exceeds
them at coarse and moderate overlap, while at strict overlap its fixed-area mask
covers only about half of the part (App.~\ref{app:strict}). Within the LLaVA family, scale yields a consistent gain
($0.43{\to}0.46$ Point on SPG, $0.53{\to}0.61$ on FS-COCO), yet within
DeepSeek-VL the larger 7B variant underperforms the 1.3B on objects
($0.23$ vs $0.31$ Point), indicating that grounding quality is not a
monotonic function of parameter count but depends on training-data
composition and instruction-tuning recipe. Fig.~\ref{fig:summary}
(right) makes the input-side dependency explicit: scene-grounding
Point on FS-COCO rises from $0.32$ at one object to $0.55$ at seven
objects, and a similar though smaller trend ($0.34{\to}0.42$ from two
to four parts) holds within SPG. Richer compositional context supplies
the cross-attention with additional disambiguating evidence, sharpening
the localisation peak. The SPG trend is the cleaner signal, since each
query targets a single part whose mask area does not grow with part
count, ruling out a coverage artefact. This pattern mirrors the pilot
(Fig.~\ref{fig:pilot}, right), where increasing abstraction along the
natural-photo $\to$ grayscale $\to$ sketch axis monotonically degraded
grounding: at one end, stripping context erodes accuracy, and at the other,
denser composition restores it. LVLM grounding therefore behaves as a
graded function of input richness, not a brittle binary of in- versus
out-of-domain.

\begin{figure}[t]
    \centering
    \includegraphics[width=.85\linewidth]{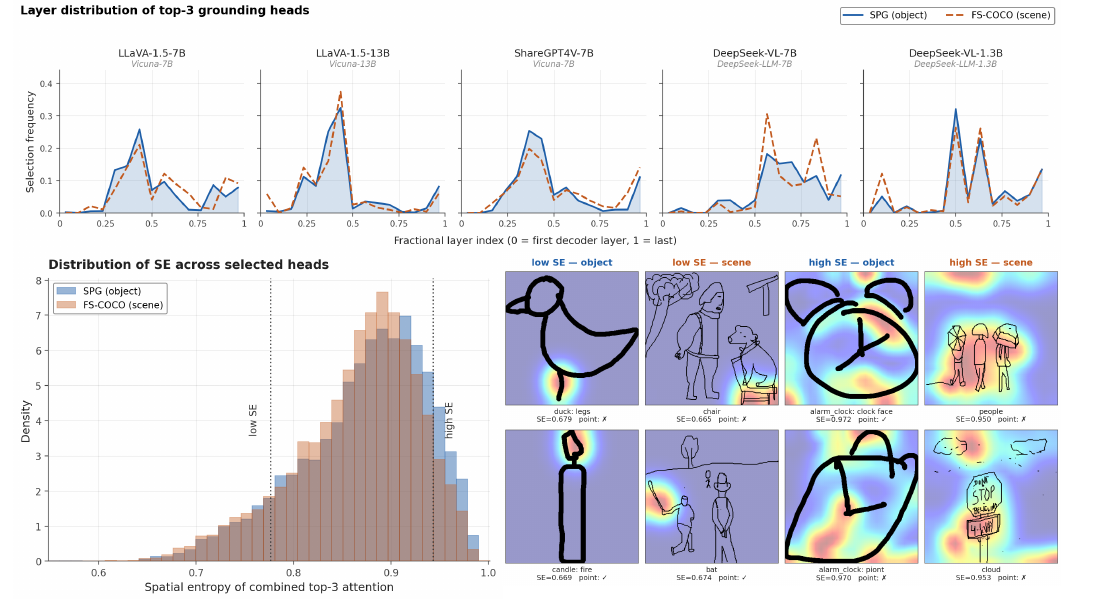}
    \caption{\textbf{Mechanistic profile of LVLM grounding.}
    \textbf{Top}: layer distribution of the top-3 grounding heads across the
    five models, SPG (object) vs FS-COCO (scene).
    \textbf{Bottom-left}: distribution of combined top-3 spatial entropy,
    pooled across models and datasets, with low/high SE tails marked.
    \textbf{Bottom-right}: LLaVA-1.5-13B attention overlays on samples drawn
    from each SE tail.}
    \label{fig:mechanistic}
\end{figure}

\textbf{Mechanistic profile.} Having established that frozen cross-attention recovers part regions, we now ask \emph{where} in the network this signal lives. Fig.~\ref{fig:mechanistic} (top) shows that the top-3 grounding heads concentrate in a narrow mid-network band in every model, but the band's location is not uniform: the three Vicuna-based models (LLaVA-1.5-7B/13B, ShareGPT4V-7B) place selection mass around fractional depth $0.35$ to $0.45$, whereas the two DeepSeek-based models sit visibly deeper, with peaks in $0.50$ to $0.65$. Crucially, the band is conserved when the input shifts from object (SPG) to scene (FS-COCO): the across-model mean shift is essentially zero ($\Delta\!=\!+0.004$, $|\Delta|\!=\!0.034$ in fractional depth), so the same layers carry grounding signal across domains. The same heads are selected in both domains, not only the same depths: on average the five models share $5.4$ of their top-10 heads, against fewer than one expected by chance, and selection frequencies correlate at $r=+0.74$ (App.~\ref{app:overlap}). Along the pilot's abstraction axis, by contrast, the peak moves earlier (Fig.~\ref{fig:pilot}), so the band is conserved across domains, not across levels of abstraction. The family split is the dominant axis of variation, not the input domain. Fig.~\ref{fig:mechanistic} (bottom-left) characterises the signal carried by these heads: their combined-top-3 spatial entropy concentrates between $0.78$ and $0.95$, with overlapping SPG and FS-COCO distributions, indicating that domain shifts the band's content as little as it shifts its location. The qualitative samples in Fig.~\ref{fig:mechanistic} (bottom-right) show what the SE tails correspond to spatially: low-SE samples tend to produce a single tightly-localised peak, whereas high-SE samples spread attention diffusely across the sketch with a correspondingly higher rate of pointing misses. SE thus measures whether the model produced a confident peak, while pointing accuracy measures whether that peak was correctly placed, and the two are complementary diagnostics rather than a single accuracy signal. Together, the three readouts portray grounding as a property of a stable, narrow layer band whose location follows the base language model in the two families we examine, while its signal quality varies with input.

\noindent\textbf{Per-class temporal-spatial coherence.}
Tab.~\ref{tab:part-ordering} and~\ref{tab:grounding} both evaluate on SPG
but probe different competences. We ask whether classes where the model
agrees with humans on part-construction order are also classes the model
grounds well, i.e., whether internalised convention (temporal) and
localisation (spatial) are coupled at the category level. For each LLaVA
variant we plot per-class LCS against per-class mAcc@0.1 across all
twenty-five SPG categories (Fig.~\ref{fig:perclass}). The two probes
dissociate in LLaVA-1.5-7B ($r=-0.13$, $\rho=-0.16$, n.s.) but couple in
LLaVA-1.5-13B ($r=+0.40$, $\rho=+0.42$, $p<0.05$): canonical-consensus
categories (\textit{suitcase}, \textit{airplane}, \textit{coffee\_cup},
\textit{butterfly}) cluster top-right, whilst ambiguous categories
(\textit{flower}) and outliers \textit{drill} (high temporal, low spatial)
and \textit{candle} (mid temporal, low spatial) occupy the off-diagonal.
LLaVA-1.5-7B orders parts at the level of a random permutation, so there is
no ordering signal to couple with, whereas in LLaVA-1.5-13B the readouts
align class by class, a diagnostic neither probe alone provides.

\subsection{The layer band is predictive, not merely descriptive}
\label{sec:additional}

\begin{wraptable}{r}{0.5\textwidth}
\centering
\small
\vspace{-4.5mm}
\setlength{\tabcolsep}{2.5pt}
\resizebox{\linewidth}{!}{\begin{tabular}{lcc}
\multirow{2}{*}{\textbf{Model}} & \textbf{SPG} \cite{li2018spg}                    & \textbf{FS-COCO} \cite{chowdhury2022fscoco}                  \\
 & \textbf{$\Delta$} (full $\rightarrow$ band)                    & \textbf{$\Delta$} (full $\rightarrow$ band)                \\
\midrule
LLaVA-1.5-7B      & +0.08 \tiny{(0.65 $\rightarrow$ \textbf{0.73})}   & +0.09 \tiny{(0.69 $\rightarrow$ \textbf{0.78})}    \\
LLaVA-1.5-13B     & +0.01 \tiny{(0.72 $\rightarrow$ \textbf{0.73})}   & +0.03 \tiny{(0.78 $\rightarrow$ \textbf{0.81})}    \\
ShareGPT4V-7B    & +0.04 \tiny{(0.73 $\rightarrow$ \textbf{0.76})}   & +0.05 \tiny{(0.73 $\rightarrow$ \textbf{0.78})}    \\
DeepSeek-VL-1.3B    & +0.04 \tiny{(0.56 $\rightarrow$ \textbf{0.60})}   & +0.04 \tiny{(0.61 $\rightarrow$ \textbf{0.65})}    \\
DeepSeek-VL-7B      & +0.01 \tiny{(0.48 $\rightarrow$ \textbf{0.49})}   & +0.04 \tiny{(0.65 $\rightarrow$ \textbf{0.69})}    \\
\end{tabular}}
\vspace{-3mm}
\end{wraptable}

\textbf{Layer band as a head-selection prior.}
We test whether the mid-network band of Fig.~\ref{fig:mechanistic} is predictive, not just descriptive. Restricting head selection to each family's empirical band ($[0.35, 0.45]$ Vicuna, $[0.50, 0.65]$ DeepSeek) matches or exceeds full-grid scoring in every (model, dataset) cell, with the largest gains on smaller models (LLaVA-1.5-7B: $+0.08$/$+0.09$ mAcc@0.1 on SPG/FS-COCO). High-scoring heads outside the modal range generalise less reliably, and the band acts as a noise-reducing prior, helping most where the model is weakest.

\textbf{The selected heads carry the signal.} Head selection is correlational, so we intervene, zeroing head outputs in the forward pass of LLaVA-1.5-13B. To avoid reading grounding off the ablated heads, we measure a downstream behaviour: whether the model can say on which side of the sketch a named part lies. Ablating the three most frequently selected heads lowers the logit margin towards the correct side by $0.118$ (95\% CI $0.074$ to $0.163$, $p=0.0001$, clustered over 24 categories), against $0.001$ for three random heads and $0.010$ for three others in the same layers. Side accuracy barely moves (App.~\ref{app:ablation}), but the layer-matched control ties the effect to the selected heads, not to their depth.

\begin{figure}[t]
  \centering
  \begin{minipage}[t]{0.49\linewidth}
    \centering
    \includegraphics[width=\linewidth]{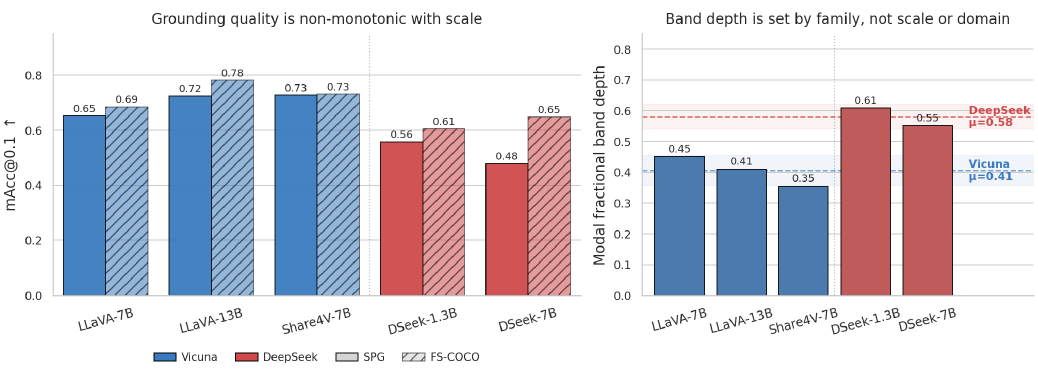}
    \caption{\textbf{Parameter count vs model family.} Left: mAcc@0.1 per (model, dataset). Right: modal fractional band depth per model, with family means (dashed) $\pm$ SD (shaded). Quality moves with scale, but band location does not.}
\label{fig:scale-band}
  \end{minipage}%
  \hfill
  \begin{minipage}[t]{0.49\linewidth}
    \centering
    \includegraphics[width=\linewidth]{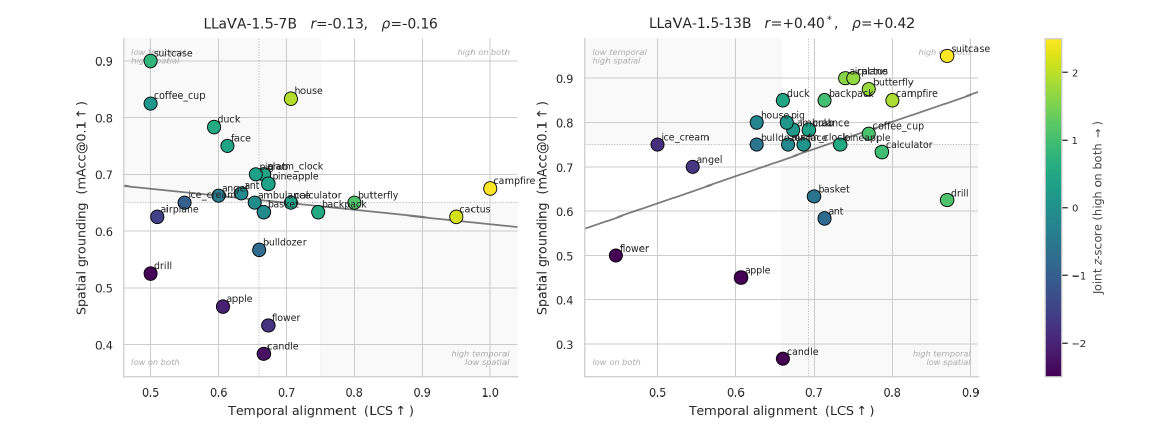}
    \caption{\textbf{Per-class temporal-spatial coherence on SPG.} LCS vs mAcc@0.1 across 25 categories, coloured by joint $z$-score. Probes dissociate in LLaVA-1.5-7B ($r=-0.13$) and couple in LLaVA-1.5-13B ($r=+0.40$, $p<0.05$).}
\label{fig:perclass}
  \end{minipage}%
  \vspace{-1em}
\end{figure}

\textbf{Within- and across-family scaling.}
Across the five models, scale and grounding quality
interact non-monotonically (Fig.~\ref{fig:scale-band}, left). The Vicuna pair improves with size
(LLaVA-1.5-7B $\to$ 13B: $+0.037$/$+0.084$ Point and $+0.072$/$+0.096$
mAcc@0.1 on SPG/FS-COCO), yet the DeepSeek pair reverses on objects
($-0.077$/$-0.078$ on SPG), with the smaller variant outperforming
the larger. Crucially, scale and domain leave the
modal grounding layer untouched (Fig.~\ref{fig:scale-band}, right): within-family SD is $0.049$ for the Vicuna trio and $0.040$ for the DeepSeek pair, whilst the cross-family gap is $0.175$, i.e., a $\sim$3.5$\times$ ratio. Per-model
modal layers are identical to three decimal places across SPG and FS-COCO.
In the two families we examine, the grounding band is therefore a property
of the base LM family, not of parameter count or input domain, extending
the split of Fig.~\ref{fig:mechanistic} to individual models.


\section{Discussion}
\label{sec:discussion}

\textbf{What the readouts show that a score does not.} Three observations here are invisible to an aggregate benchmark. First, model differences are structured: DeepSeek-VL-7B trails DeepSeek-VL-1.3B on object grounding, yet both place their grounding heads in the same band (Sec.~\ref{sec:additional}). Second, failure is graded: under abstraction the model becomes wrong about \emph{where} a part lies before its attention becomes diffuse (Sec.~\ref{sec:method_motivation}). Third, models defer to human convention: agreement rises and falls with human consensus, so variability in ambiguous categories is informative, not simply a lower score.

\textbf{Limitations.} The probes assume the queried part is present. Asked about absent parts taken from other categories, the model never signals absence and answers as it does for real parts, as in \emph{``The bell of the airplane is located at the top of the wing''} and \emph{``The stamens of the alarm clock are located at the top of the clock''} (App.~\ref{app:phantom}). The readout thus says \emph{where} a named part is, not \emph{whether} it is there, and pipelines in which parts may be absent need a separate presence check. The attention readouts also cover only two base language model families, and further families are future work.

\section{Conclusion}
We used sketches as a controlled abstraction test for LVLMs: strip texture and context, keep human-selected structure, and ask what survives. The resulting training-free probe yields three readouts: part-construction order tracks human convention where stable and is carried by the language model, frozen attention heads recover part regions without grounding supervision, and a narrow mid-network band set by the base language model in the families we examine acts as a head-selection prior that improves grounding without retraining and couples with the temporal readout at scale. Sketches thus become a diagnostic input rather than a domain-shift benchmark, exposing what LVLMs retain, when it becomes unreliable, and which layers carry it.

\clearpage
\section*{Acknowledgements}
We acknowledge the support of UK Research and Innovation (UKRI) through the UKRI AI: Centre for Doctoral Training in AI for Digital Media Inclusion under Grant No.: EP/Y030915/1. For the purpose of open access, the author has applied a Creative Commons Attribution (CC BY) license to any Author Accepted Manuscript version arising.

{\small
\bibliographystyle{plain}
\bibliography{ref}
}





\clearpage
\appendix
\etocdepthtag.toc{mtappendix}
\etocsettagdepth{mtmain}{none}
\etocsettagdepth{mtappendix}{subsection}
\etocsettocstyle{\noindent{\Large\bfseries Appendix}\par\vspace{1.5em}\noindent{\large\bfseries Appendix Contents}\par\vspace{0.75em}}{}
\tableofcontents
\clearpage

\section{Implementation details}
\label{app:impl}

\subsection{Temporal probe prompt}
\label{app:prompt}
The temporal probe uses the system and user prompt below, reproduced verbatim. The part names are inserted in a shuffled order that is fixed per sketch, as a quoted, comma-separated list in set notation, so the listing carries no ordering cue. The response is parsed for the \texttt{order} field, and each returned name is matched to the given parts after lower-casing and normalising underscores and hyphens.

\begin{tcolorbox}[colback=Gray,colframe=Gray,boxrule=0pt,arc=2pt,left=5pt,right=5pt,top=4pt,bottom=4pt]
\small\ttfamily\frenchspacing\raggedright
\textbf{System:} You analyse hand-drawn sketches and infer the temporal order in which a human drew their parts. Your inference is based purely on visual evidence in the sketch \textemdash{} relative size, position, layering, and which parts visually rest on or extend from others \textemdash{} combined with general knowledge of how humans typically construct drawings.\par\medskip
\textbf{User:} You are looking at a black-line sketch of a \{class\}.\par\smallskip
The sketch contains the following parts (presented as an unordered set; the listing order has no relation to the drawing order):\par
\$\textbackslash\{"\{part\}", "\{part\}", \ldots\textbackslash\}\$\par\smallskip
Question: in what order did the human draw these parts?\par\smallskip
Output a single JSON object with two fields:\par
- "reason": one sentence describing the visual evidence you used.\par
- "order": the parts arranged from earliest drawn to latest drawn.\par\smallskip
Constraints:\par
- Use only parts from the set; do not invent or modify part names.\par
- Use each part exactly once.\par\smallskip
Example format:\par
\{"reason": "...", "order": ["first", "second", ...]\}
\end{tcolorbox}

\subsection{Grounding pipeline}
\label{app:grounding}
For each query we read the attention from the last input token to the image patches in every head (Sec.~\ref{sec:method_main}). Heads are scored by image-attention mass and spatial entropy, the candidates are cut with the chord-elbow rule of Kang et al.~\cite{kang2025localisationheads}, and the top $k=3$ are kept. Each selected map is mean-subtracted, passed through a ReLU, smoothed with a Gaussian of $\sigma=1$, and normalised. The three maps are combined by element-wise maximum and binarised at the 80th percentile, so the predicted mask always covers the top 20\% of the patch grid ($24\times24$ for the LLaVA-1.5 models). The baselines binarise differently: CLIPSeg~\cite{lueddecke2022clipseg} thresholds its sigmoid output at $0.5$, and GroupViT~\cite{xu2022groupvit} assigns each pixel to its highest-scoring group. A fixed-area mask is favoured at a loose IoU threshold and penalised at a strict one, which App.~\ref{app:strict} examines. SPG ground truth for a part is the convex hull of its strokes, and FS-COCO ground truth is the object mask from FS-COCO-Seg~\cite{bourouis2024scenesketch}.

\subsection{Compute}
All experiments are forward passes of frozen models on a single NVIDIA A100 GPU. Closed-source models are queried through their vendor APIs. No model is trained or fine-tuned.

\section{Human consensus and LVLM agreement}
\label{app:consensus}
Human consensus for a category measures how consistently SPG drawings follow its modal part order, as ranked in Fig.~\ref{fig:spg_consensus}. Tab.~\ref{tab:consensus} correlates it with each model's per-category LCS over the 25 categories. Memorisation and retained convention make different predictions here. A model reproducing stored SPG trajectories would agree equally well whether or not humans converge, whereas a model that has retained the convention should agree more where the convention is stronger. Eight of the nine models show a significant positive correlation. In those eight, agreement on high-consensus categories exceeds that on low-consensus categories by $0.09$ to $0.24$ LCS. The exception is LLaVA-1.5-7B, whose agreement is at the level a random permutation of the same parts would achieve, so there is no ordering signal for consensus to track.

An independent check uses the model's own consensus, the agreement among its predictions across the sketches of a category. It also correlates with human consensus, in all nine models ($r=+0.35$ to $+0.79$), significantly in eight. With 25 categories both analyses are modestly powered, and we read them as evidence rather than proof.

\begin{table}[h]
  \centering
  \small
  \caption{\textbf{Human consensus against LVLM agreement.} Correlation between per-category human consensus (Fig.~\ref{fig:spg_consensus}) and per-category LCS with the human trajectory, over the 25 SPG categories.}
  \label{tab:consensus}
  \begin{tabular}{lccc}
    \toprule
    \textbf{Model} & \textbf{Pearson $r$} & \textbf{$p$} & \textbf{Spearman $\rho$} \\
    \midrule
    Gemini 2.5 Flash-Lite~\cite{comanici2025gemini25} & $+0.82$ & $<0.001$ & $+0.77$ \\
    Claude Haiku 4.5~\cite{anthropic2025claude45}     & $+0.77$ & $<0.001$ & $+0.80$ \\
    Qwen2.5-VL-7B~\cite{qwen25vl2025}                 & $+0.71$ & $<0.001$ & $+0.66$ \\
    Molmo-7B-D~\cite{deitke2025molmo}                 & $+0.70$ & $<0.001$ & $+0.64$ \\
    GPT-5 (nano)~\cite{openai2025gpt5}                & $+0.61$ & $0.001$  & $+0.64$ \\
    InternVL2-8B~\cite{chen2024internvl}              & $+0.60$ & $0.002$  & $+0.59$ \\
    LLaVA-1.5-13B~\cite{liu2024llava}                 & $+0.54$ & $0.005$  & $+0.59$ \\
    Qwen2.5-VL-3B~\cite{qwen25vl2025}                 & $+0.54$ & $0.006$  & $+0.48$ \\
    LLaVA-1.5-7B~\cite{liu2024llava}                  & $-0.06$ & n.s.     & $-0.25$ \\
    \midrule
    Mean over nine models                             & $+0.58$ &          &         \\
    \bottomrule
  \end{tabular}
\end{table}

\section{Image controls for the temporal probe}
\label{app:controls}

\paragraph{Protocol.} The controls use 50 sketches per category and the prompt of App.~\ref{app:prompt}. Each sketch has one shuffle of its parts, drawn once with a fixed seed and reused in every condition, and the object name in the prompt is always the true category. \emph{Blank canvas} replaces the sketch with a $512\times512$ white image. \emph{No image} sends the prompt alone, with no image tokens. \emph{Mismatched category} shows a sketch from the category whose part vocabulary overlaps least with the target's. Parts that a response omits are appended in their shuffled input order, so a response that cannot be parsed scores at the level of a random ordering rather than of the reference. Each control is compared with the real sketch on per-category means by a Wilcoxon signed-rank test over the 25 categories. Two models, three controls, and three metrics give the eighteen comparisons of Sec.~\ref{sec:main_results}. None favours the real sketch, and the only significant one favours the no-image condition for Qwen2.5-VL-7B on Pair ($p=0.04$).

\paragraph{Deviation set.} The main protocol samples sketches whose drawing order matches the category's modal order, so every sketch in a category has the same correct answer, which the text alone can supply. The deviation set instead takes sketches with the same parts as the modal order but a different drawing order. The prompt is then identical to the main protocol and only the image, and with it the correct answer, changes. We test whether the prediction is closer to the sketch's own order than the modal order is. A per-sketch shift in that direction does not survive clustering by category ($p=0.26$, 17 categories).

\paragraph{Forced choice.} The model sees one sketch and two candidate orderings, its own and another, matched for edit distance from the modal order, and must say which belongs to the sketch. A model that reasons from drawing conventions alone cannot prefer either, so chance is 50\% by construction. Qwen2.5-VL-7B is at chance both with the sketch and without it.

\section{Grounding at stricter overlap}
\label{app:strict}
Tab.~\ref{tab:strict_full} re-scores every method at three IoU thresholds from the stored per-part IoU, with no new inference. Two mAcc@0.1 entries differ from Tab.~\ref{tab:grounding} by $0.01$, and we report this run unchanged so that no row mixes runs. Grounding is evaluated on the same sketches for every method, so we pair the comparison by category with a Wilcoxon signed-rank test (25 categories for SPG, 185 for FS-COCO). At mAcc@0.1 LLaVA-1.5-13B is above both trained grounders on both datasets, and all four comparisons are significant ($p<0.002$). At mAcc@0.3 it remains above both, significantly in three of four comparisons. At mAcc@0.5 CLIPSeg is nominally ahead on both datasets and GroupViT on SPG, and no difference is significant.

\begin{table}[h]
  \centering
  \small
  \caption{\textbf{Grounding at three overlap thresholds.} All methods re-scored from one run.}
  \label{tab:strict_full}
  \begin{tabular}{l ccc ccc}
    \toprule
    & \multicolumn{3}{c}{\textbf{SPG~\cite{li2018spg}}} & \multicolumn{3}{c}{\textbf{FS-COCO~\cite{chowdhury2022fscoco}}} \\
    \cmidrule(lr){2-4}\cmidrule(lr){5-7}
    \textbf{Model} & \textbf{@0.1$\uparrow$} & \textbf{@0.3$\uparrow$} & \textbf{@0.5$\uparrow$} & \textbf{@0.1$\uparrow$} & \textbf{@0.3$\uparrow$} & \textbf{@0.5$\uparrow$} \\
    \midrule
    CLIPSeg~\cite{lueddecke2022clipseg} & 0.27 & 0.17 & 0.12 & 0.40 & 0.31 & 0.20 \\
    GroupViT~\cite{xu2022groupvit}      & 0.58 & 0.28 & 0.11 & 0.60 & 0.21 & 0.08 \\
    LLaVA-1.5-13B~\cite{liu2024llava}   & 0.72 & 0.29 & 0.06 & 0.78 & 0.35 & 0.11 \\
    \bottomrule
  \end{tabular}
\end{table}

\paragraph{Threshold sweep.} To locate the cause of the drop at strict overlap, we sweep the binarisation percentile over $\{60,70,75,80,85,90,95\}$ for LLaVA-1.5-13B on SPG. At the 80th percentile used throughout, the mask has precision $0.35$ and recall $0.54$ and covers about $1.5$ times the ground-truth area. mAcc@0.5 rises as the percentile is lowered, that is, as the mask grows. The shortfall at strict overlap therefore comes from under-coverage at a fixed threshold rather than from an over-large mask. Two properties of the evaluation bear on all thresholds. The LVLM mask has a fixed area while CLIPSeg's adapts to the part (App.~\ref{app:grounding}), and the convex-hull ground truth of SPG is larger and smoother than the inked region, which favours diffuse predictions.

\section{Object and scene head overlap}
\label{app:overlap}
For each model and dataset we count how often each (layer, head) cell enters the top-3 selection across all grounding queries. Tab.~\ref{tab:overlap} compares SPG (object) with FS-COCO (scene) by the modal fractional depth of these counts, by the number of the ten most frequently selected heads that the two datasets share, and by the Pearson correlation of selection frequencies over all cells. The datasets have different queries, so the comparison is between distributions rather than paired queries. Two unrelated top-10 sets drawn from a grid of $L\times H$ cells share $100/(LH)$ heads in expectation, well under one for every model here. The modal depth is identical to three decimal places in every model, and on average $5.4$ of the top ten heads are shared.

\begin{table}[h]
  \centering
  \small
  \caption{\textbf{Grounding-head overlap between object and scene sketches.} Depth is the modal fractional depth of top-3 selections.}
  \label{tab:overlap}
  \begin{tabular}{lcccc}
    \toprule
    \textbf{Model} & \textbf{Depth, SPG} & \textbf{Depth, FS-COCO} & \textbf{Shared top-10} & \textbf{Frequency $r$} \\
    \midrule
    ShareGPT4V-7B~\cite{chen2024sharegpt4v}  & 0.355 & 0.355 & 6/10 & $+0.72$ \\
    LLaVA-1.5-13B~\cite{liu2024llava}        & 0.410 & 0.410 & 5/10 & $+0.74$ \\
    LLaVA-1.5-7B~\cite{liu2024llava}         & 0.452 & 0.452 & 5/10 & $+0.66$ \\
    DeepSeek-VL-7B~\cite{lu2024deepseekvl}   & 0.552 & 0.552 & 4/10 & $+0.70$ \\
    DeepSeek-VL-1.3B~\cite{lu2024deepseekvl} & 0.609 & 0.609 & 7/10 & $+0.87$ \\
    \midrule
    Mean                                     &       &       & 5.4/10 & $+0.74$ \\
    \bottomrule
  \end{tabular}
\end{table}

\section{Head ablation}
\label{app:ablation}
We ablate a head by zeroing its slice of the input to the attention output projection, which removes its contribution to the residual stream, in every forward pass of LLaVA-1.5-13B. Each query asks \emph{``Is the \{part\} of the \{category\} on the left or the right side of this sketch? Answer with one word.''} The correct answer is the side on which the centroid of the part's ground-truth mask falls, and parts whose centroid lies within $0.12$ of the image width from the midline are excluded, so a correct answer requires localisation. We read the next-token logits rather than generating, take the maximum logit over the tokens for \emph{left} and for \emph{right} (with and without capitalisation and a leading space), and define the margin as the correct side's logit minus the other's.

The top-3 heads are the three most frequently selected across the SPG grounding queries of LLaVA-1.5-13B. Random heads are drawn uniformly from the rest of the grid, and layer-matched heads from the other heads in the layers of the top-3, both with a fixed seed. Per-category mean margins are compared with the unablated model by a Wilcoxon signed-rank test over the 24 categories with queries in every condition, with a 95\% bootstrap interval over categories. Paired over categories, the top-3 drop also exceeds the random ($p=0.0002$) and layer-matched ($p=0.0003$) drops. The majority-class rate for side accuracy is $0.586$. Accuracy barely moves under any ablation (Tab.~\ref{tab:ablation}), since three of 1{,}600 heads are too few to flip many decisions, while the margin shows that the selected heads, and not other heads at the same depth, carry the signal.

\begin{table}[h]
  \centering
  \small
  \caption{\textbf{Head ablation in LLaVA-1.5-13B.} Side accuracy and logit margin towards the correct side.}
  \label{tab:ablation}
  \begin{tabular}{lccc}
    \toprule
    \textbf{Ablated heads} & \textbf{Side accuracy$\uparrow$} & \textbf{Margin$\uparrow$} & \textbf{Change in margin} \\
    \midrule
    None                      & 0.701 & 0.854 & \\
    Top-3 grounding heads     & 0.713 & 0.736 & $\mathbf{-0.118}$ ($p=0.0001$) \\
    Random 3 heads            & 0.701 & 0.854 & $-0.001$ \\
    3 heads, same layers      & 0.695 & 0.844 & $-0.010$ \\
    \bottomrule
  \end{tabular}
\end{table}

\section{Phantom parts}
\label{app:phantom}
To test what the probe does when there is nothing to find, we query parts that are absent from the sketch. For each SPG category we draw two part names from the vocabularies of other categories, excluding any name that occurs in the target category under any spelling, and query them on ten sketches per category with the spatial template of Sec.~\ref{sec:method_main}, alongside the real parts of the same sketches. We also pose the same question for a text answer. The model never states that a part is absent, and its answers about phantom parts read like its answers about real ones:
\begin{itemize}
  \item phantom: \emph{``The bell of the airplane is located at the top of the wing.''}
  \item phantom: \emph{``The stamens of the alarm clock are located at the top of the clock.''}
  \item real: \emph{``The body of the airplane is located in the middle of the drawing.''}
\end{itemize}
The probe therefore localises without verifying existence. It answers where a named part is, not whether it is there, and a pipeline in which parts may be absent needs a separate presence check.
\section{Newer open models}
\label{app:qwen3}
The temporal probe needs only generated text, so it extends directly to newer releases. We ran Qwen3-VL~\cite{qwen3vl2025} at 4B and 8B under the identical protocol, which gives a generational comparison against the Qwen2.5-VL rows of Tab.~\ref{tab:part-ordering} with family and probe held fixed (Tab.~\ref{tab:qwen3}). The generational gain is large at the smaller scale ($+0.16$ Edit and $+0.20$ Pair from Qwen2.5-VL-3B to Qwen3-VL-4B) and small at the larger ($+0.03$ and $+0.03$ from 7B to 8B). The two Qwen3-VL sizes are almost indistinguishable, so agreement has saturated within the family. Qwen3-VL-4B matches or exceeds GPT-5 (nano) and Gemini 2.5 Flash-Lite on all three metrics, and the gap to Claude Haiku 4.5 narrows to $0.04$ LCS. In line with App.~\ref{app:controls}, these numbers measure the drawing convention the language model has retained.

\begin{table}[h]
  \centering
  \small
  \caption{\textbf{Part-construction order for Qwen3-VL} on SPG, protocol as in Tab.~\ref{tab:part-ordering}.}
  \label{tab:qwen3}
  \begin{tabular}{lccc}
    \toprule
    \textbf{Model} & \textbf{Edit$\uparrow$} & \textbf{LCS$\uparrow$} & \textbf{Pair$\uparrow$} \\
    \midrule
    Qwen2.5-VL-3B (Tab.~\ref{tab:part-ordering}) & 0.57 & 0.77 & 0.64 \\
    Qwen3-VL-4B                                   & \textbf{0.73} & \textbf{0.85} & \textbf{0.84} \\
    \midrule
    Qwen2.5-VL-7B (Tab.~\ref{tab:part-ordering}) & 0.69 & 0.84 & 0.80 \\
    Qwen3-VL-8B                                   & \textbf{0.72} & \textbf{0.85} & \textbf{0.83} \\
    \midrule
    Claude Haiku 4.5 (Tab.~\ref{tab:part-ordering}) & 0.80 & 0.89 & 0.88 \\
    \bottomrule
  \end{tabular}
\end{table}


\end{document}